\documentclass[11pt,a4paper]{article}
\usepackage[hyperref]{emnlp2018}
\usepackage{times}
\usepackage{latexsym}

\usepackage{url}

\usepackage{times}
\usepackage{xcolor}
\usepackage{natbib}
\usepackage{soul}
\usepackage[utf8]{inputenc}
\usepackage[small]{caption}

\usepackage{CJKutf8}
\usepackage{hyperref}
\usepackage{url}

\usepackage{amsmath,amsthm,amssymb,amsfonts,bbm}
\usepackage{enumitem}

\usepackage{multirow} 
\usepackage{graphicx} 
\usepackage{subfigure}
\usepackage{booktabs} 
\usepackage[normalem]{ulem}
\usepackage{makecell}
\usepackage{courier}
\usepackage{bbm}
\usepackage[noend]{algpseudocode}
\usepackage{algorithmicx,algorithm}
\usepackage{adjustbox}
\usepackage{tabularx}
\usepackage{booktabs}
\usepackage{setspace}
\usepackage{threeparttable}
\usepackage[ruled,boxed,algo2e]{algorithm2e}
\usepackage{algorithm}
\usepackage{listings}
\usepackage{pifont}
\usepackage{color}
\usepackage{bm}
\usepackage{array}

\usepackage{arydshln}
\newcommand{\mb}{\mathbf}
\newcommand{\x}{\mathbf{x}}
\newcommand{\y}{\mathbf{y}}
\newcommand{\z}{\mathbf{z}}
\newcommand{\X}{\mathbf{X}}

\newcommand{\txtitcolor}[2]{\textcolor{#1}{\textit{#2}}}

\aclfinalcopy 

\title{Language Style Transfer from Sentences with Arbitrary Unknown Styles}

\author{
  $\text{Yanpeng Zhao}^1$\thanks{\,\,\,This work was done while Yanpeng Zhao was with Tencent AI Lab.}, $\text{Wei Bi}^2$\thanks{\,\,\,Corresponding author.}, $\text{Deng Cai}$, $\text{Xiaojiang Liu}^2$, $\text{Kewei Tu}^1$, $\text{Shuming Shi}^2$ \\
  $^1\text{ShanghaiTech University, Shanghai, China}$ \\
  $^2\text{Tencent AI Lab, Shenzhen, China}$\\
  {\tt \{zhaoyp1,tukw\}@shanghaitech.edu.cn},
  {\tt {\tt thisisjcykcd@gmail.com}} \\
  {\tt \{victoriabi,kieranliu,shumingshi\}@tencent.com}\\
  }

\date{}

\begin{document}
\maketitle
\begin{abstract}
Language style transfer is the problem of migrating the content of a source sentence to a target style.
In many of its applications, parallel training data are not available and source sentences to be transferred may have arbitrary and unknown styles.
Under this problem setting, we propose an encoder-decoder framework.
First, each sentence is encoded into its content and style latent representations.
Then, by recombining the content with the target style, we decode a sentence aligned in the target domain.
To adequately constrain the encoding and decoding functions, we couple them with two loss functions.
The first is a style discrepancy loss, enforcing that the style representation accurately encodes the style information guided by the discrepancy between the sentence style and the target style.
The second is a cycle consistency loss, which ensures that the transferred sentence should preserve the content of the original sentence disentangled from its style.
We validate the effectiveness of our model in three tasks: sentiment modification of restaurant reviews, dialog response revision with a romantic style, and sentence rewriting with a Shakespearean style.
\end{abstract}

\section{Introduction}

Style transfer aims at migrating the content of a sample from a source style to a target style.
Recently, great progress has been achieved by applying deep neural networks to redraw an image in a particular style~\citep{kulkarni2015deep,liu2016coupled,gatys2016image,zhu2017unpaired,luan2017deep}.
However, so far very few approaches have been proposed
for style transfer of natural language sentences,
i.e., changing the style or genre of a sentence while preserving its semantic content.
For example, we would like a system that can convert a given text piece in the language of Shakespeare~\citep{mueller2017sequence};
or rewrite product reviews with a favored sentiment~\citep{shen2017style};
or generate responses with a consistent persona~\cite{li2016persona}.

\begin{table}
	{\setlength{\tabcolsep}{.6em}
		\makebox[\linewidth]{\resizebox{\linewidth}{!}{%
	\begin{tabular}{>{\raggedleft\arraybackslash}p{0.13\textwidth}l}
	\toprule
	\txtitcolor{blue}{user} & Have you ever been to Moribor before? \\
	\txtitcolor{black}{bot} & Yes, I have been to Moribor twice. \\
	\txtitcolor{red}{transferred} & Yes, I have been to Moribor twice, it is beautiful.\\
	\hdashline
	\txtitcolor{blue}{user} & How do you like you neighbors there?\\
	\txtitcolor{black}{bot} & They were very polite and quiet throughout the night. \\
	\txtitcolor{red}{transferred} & They were very kind and made me feel comfortable. \\
	\hdashline
	\txtitcolor{blue}{user} & What was the weather like during your stay?\\
	\txtitcolor{black}{bot} & The muggy weather made me cranky. \\
	\txtitcolor{red}{transferred} & The warm sunlight made it a great day. \\
	\hdashline
	\txtitcolor{blue}{user} & I heard you even met your father there.\\
	\txtitcolor{black}{bot} & Um, I thought he would not go there.\\
	\txtitcolor{red}{transferred} & Um, I was very glad to meet him there.\\
	\hdashline
	\txtitcolor{blue}{user} & How is he getting on?\\
	\txtitcolor{black}{bot} & Terrible! No one wants to make friends with him.\\
	\txtitcolor{red}{transferred} & Great! Everyone wants to be his friend.\\
	\bottomrule
	\end{tabular}}}}
	\caption{Chatbot responses have inconsistent sentiments. Transferred responses have a consistent positive sentiment regardless of the sentiments of the original responses (bot).}
\vspace*{-1.2em}
\end{table}

A big challenge faced by language style transfer is that large-scale parallel data are unavailable.
However, parallel data are necessary for most text generation frameworks,  
such as the popular sequence-to-sequence models~\citep{sutskever2014sequence,bahdanau2014neural,rush2015neural,nallapati2016abstractive,paulus2017deep}.
Hence these methods are not applicable to the language style transfer problem.
A few approaches have been proposed to deal with non-parallel data~\citep{hu2017toward,shen2017style}. Most of these approaches try to learn a latent representation of the content disentangled from the source style,
and then recombine it with the target style to generate the corresponding sentence.

All the above 
approaches assume that data have only two styles, and their task is to transfer sentences from one style to the other. 
However, in many practical settings, we may deal with sentences with arbitrary unknown styles.
Consider we are building chatbots.
A good chatbot needs to exhibit a consistent persona, so that it can gain the trust of users. 
However, existing chatbots such as Siri, Cortana, and XiaoIce~\citep{harry2018} lack the ability of generating responses 
with a consistent persona during the whole conversation.
Table 1 shows some examples. The chatbot responds to the user with varying sentiments (neutral, positive, or negative). 
One possible solution is to transfer the generated chatbot responses into a target persona before sending them to users. 
Hence, in this paper, we study the setting of language style transfer in which the source data to be transferred can have arbitrary unknown styles.

Another challenge in language style transfer is that the transferred sentence should preserve the content of the original sentence disentangled from its style. To tackle this problem, \citet{shen2017style} assumed the source and target domain share the same latent content space, and trained their model by aligning these two latent spaces. 
\citet{hu2017toward} constrained that the latent content representation of the original sentence could be inferred from the transferred sentence. However, these attempts considered content modification in the latent content space but not the sentence space.


The contribution of this paper mainly consists of the following three parts:
\begin{itemize}[wide=0\parindent,noitemsep]
\item[1.] We address a new style transfer task where sentences in the source domain can have arbitrary language styles but those in the target domain are with only one language style. 
\item[2.] We propose a style discrepancy loss to learn disentangled representations of content and style. 
This loss enforces that the discrepancy between an arbitrary style representation and the target style representation should be consistent with the closeness of its sentence style to the target style. 
Additionally, we employ a cycle consistency to avoid content change.
\item[3.] 
We evaluate our model in three tasks: sentiment modification of restaurant reviews, dialog response revision with a romantic style, and sentence rewriting with a Shakespearean style.
Experimental results show that our model surpasses the state-of-the-art style transfer model~\citep{shen2017style} in these three tasks.
\end{itemize}

\section{Related Work} \label{sec:related}

\noindent\textbf{Image Style Transfer:}
Most style transfer approaches in the literatures focus on vision data.
\citet{kulkarni2015deep} proposed to disentangle the content representations from image attributes, and control the image generation by manipulating the graphics code that encodes the attribute information. ~\citet{gatys2016image} used Convolutional Neural Networks (CNNs) to learn separated representations of the image content and style, and then created the new image from their combination. 
Some approaches have been proposed to align the two data domains with the idea of generative adversarial networks (GANs)~\citep{goodfellow2014generative}.
~\citet{liu2016coupled} proposed a coupled GAN framework to learn a joint distribution of multi-domain data by the weight-sharing constraint. 
~\citet{zhu2017unpaired} introduced a cycle consistency loss,
which minimizes the gap between the transferred images and the original ones. 
However, due to the discreteness of the natural language, this loss function cannot be directly applied on text data.
In our work, we show how the idea of cycle consistency can be used on text data.

\vspace*{-0.01em}
\noindent\textbf{Text Style Transfer:}
To handle the non-parallel data problem, 
\citet{mueller2017sequence} revised the latent representation of a sentence in a certain direction guided by a classifier, 
so that the decoded sentence imitates those favored by the classifier.
\citet{ficler2017controlling} encoded textual property values with embedding vectors, and adopted a conditioned language model to generate sentences satisfying the specified content and style properties.
\citet{li2018delete} demonstrated that a simple delete-retrieve-generate approach could achieve good performance in sentiment transfer tasks. 
\citet{hu2017toward} used the variational auto-encoder (VAE)~\cite{kingma2013auto} to encode the sentence into a latent content representation disentangled from the source style, and
then recombine it with the target style to generate its counterpart.
\citet{shen2017style} considered transferring between two styles simultaneously.
They utilized adversarial training 
to align the generated sentences from one style to the data domain of the other style. 
We also adopt similar adversarial training in our model. However, since we assume the source domain contains data with various and possibly unknown styles, 
it is impossible for us to apply a discriminator to determine whether a sentence transferred from the target domain is aligned in the source domain as in~\citet{shen2017style}.

\vspace*{-0.3em}
\section{Problem Formulation}

We now formally present our problem formulation.
Suppose there are two data domains,
one source domain 
$\mathcal{X}_s$
in which each sentence may have its own language style,
and one target domain 
$\mathcal{X}_t$
consisting of data with the same language style. 
During training, we observe $n$ samples from $\mathcal{X}_s$ and $m$ samples from $\mathcal{X}_t$,
denoted as 
$\mb{X}_{s} = \{\mb{x}_{s}^{(1)}, \mb{x}_{s}^{(2)}, \ldots, \mb{x}_{s}^{(n)}\}$ and
$\mb{X}_{t} = \{\mb{x}_{t}^{(1)}, \mb{x}_{t}^{(2)}, \ldots, \mb{x}_{t}^{(m)}\}$.
Note that we can hardly find a sentence pair $(\x_s^{(i)}, \x_t^{(j)})$
that describes the same content.
Our task is to design a model to learn from these non-parallel training data such that for an unseen testing sentence $\x \in \mathcal{X}_s$,
we can transfer it into its counterpart $\tilde{\x} \in \mathcal{X}_t$, where $\tilde{\x}$ should preserve the content of $\x$ but
with the language style in $\mathcal{X}_t$.

\section{Model} \label{sec:model}	

\subsection{Encoder-decoder Framework}

\begin{figure*}[tb]
\begin{center}
\subfigure[]{\label{fig:model1}
  \includegraphics[width=0.65\linewidth]{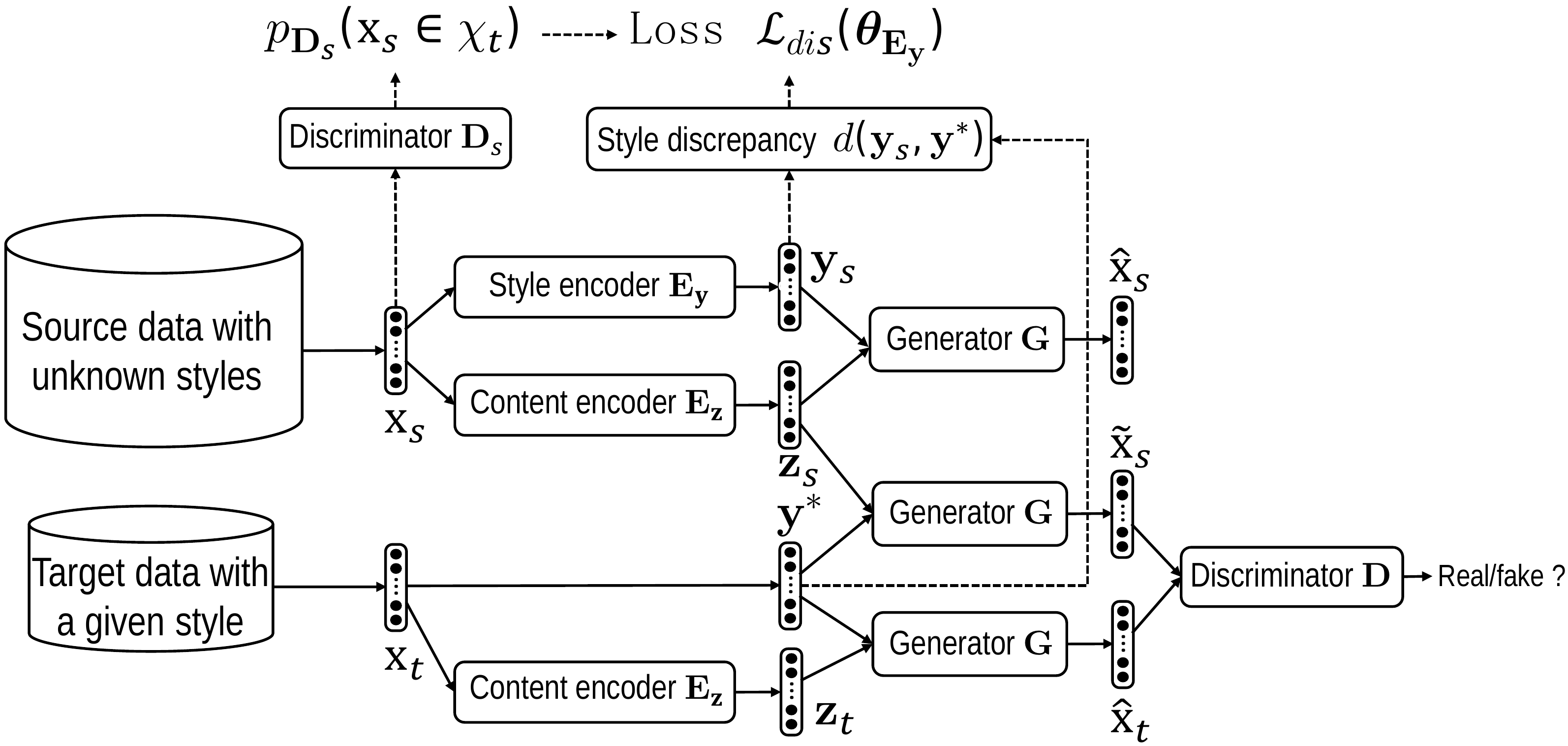}}
\subfigure[]{\label{fig:model2}
 \includegraphics[width=0.32\linewidth]{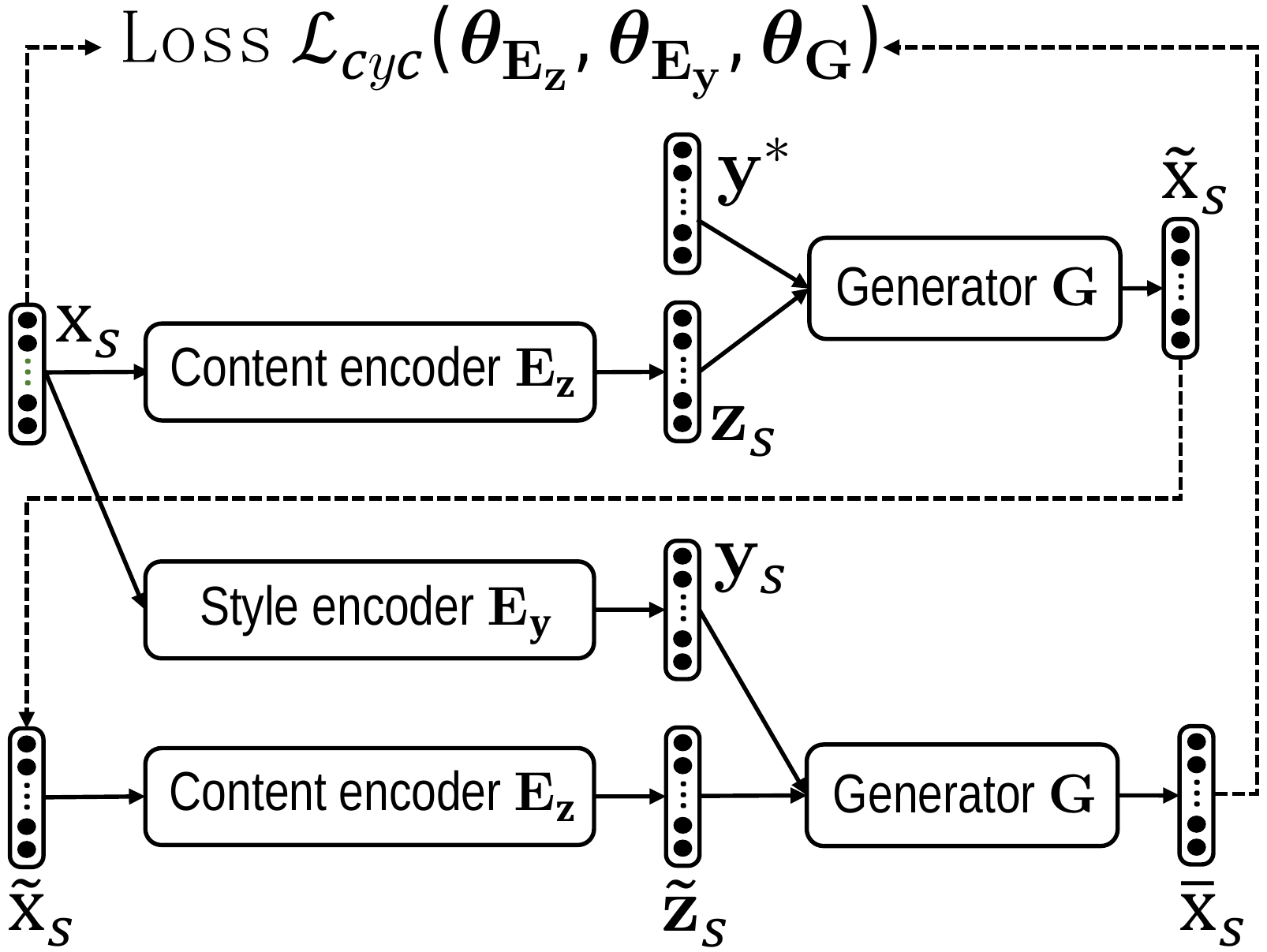}
 }
\end{center}
\vspace{-10pt}
\caption{(a) Basic model with the style discrepancy loss. Solid lines: encode and decode the sample itself; dash lines: transfer $\x_1 \in \X_1$ into $\mathcal{X}_2$. (b): Proposed cycle consistency loss (can be applied for samples in $\mathcal{X}_2$ similarly).}
\label{fig:a}
\vspace{-15pt}
\end{figure*}


We assume each sentence $\x$ can be decomposed into two representations:
one is the style representation $\y \in \mathcal{Y}$, and the other is the content representation $\z \in \mathcal{Z}$, which is disentangled from its style.
Each sentence $\x_s^{(i)} \in \X_{s}$ has its individual style $\mb{y}_{s}^{(i)}$,
while all the sentences $\x_t^{(i)} \in \X_t$  share the same style, denoted as $\mb{y}^*$.	
Our model is built upon the encoder-decoder framework. 
In the encoding module, we assume that $\z$ and $\y$ of a sentence $\x$ can be obtained through two encoding functions
$\mb{E}_{\z}(\x)$ and $\mb{E}_{\y}(\x)$ respectively:
\begin{align}
\z_s^{(i)} &= \mb{E}_{\z}(\x_s^{(i)}),  \quad \y_s^{(i)} = \mb{E}_{\y}(\x_s^{(i)})\,,\nonumber\\
\z_t^{(j)} &= \mb{E}_{\z}(\x_t^{(j)}), \quad \y^{*}\,\, = \mb{E}_{\y}(\x_t^{(j)})\,,\nonumber
\end{align}
where $i \in \{1,2,\ldots,n\}$, $j \in \{1,2,\ldots,m\}$, $\mb{E}_{\y}(\x) = \mathbbm{1}_{\{\x \in \X_s\}} \cdot g(\x) + \mathbbm{1}_{\{\x \in \X_t\}}  \cdot \y^*$, and $\mathbbm{1}_{\{\cdot\}}$ is an indicator function. When a sentence $\x$ comes from the source domain, we use a function $g(\x)$ to encode its style representation. For $\x$ from the target domain, a shared style representation $\y^*$ is used. Both $\y^*$ and parameters in $g(\x)$ are learnt jointly together with other parameters in our model.

For the decoding module, we first employ a reconstruction loss to encourage that the sentence from the decoding function given $\z$ and $\y$ of a sentence $\x$ can well reconstruct $\x$ itself.
Here, we use a probabilistic generator $\mb{G}$ as the decoding function and the reconstruction loss is:
\begin{align} \label{eq:rec}
\mathcal{L}_{rec} (& \bm{\theta}_{\mb{E}_{\z}}, \bm{\theta}_{\mb{E}_{\y}}, \bm{\theta}_{\mb{G}})\nonumber \\
=& \mathbb{E}_{\mb{x}_{s} \sim \mb{X}_{s}}\left[-\log p_{\mb{G}}(\mb{x}_{s} | \mb{y}_s, \mb{z}_{s} )\right] \nonumber \\
+& \mathbb{E}_{\mb{x}_{t} \sim \mb{X}_{t}}\left[-\log p_{\mb{G}}(\mb{x}_{t} | \mb{y}^*, \mb{z}_t)\right],
\end{align}
where $\bm{\theta}$ denotes the parameter of the corresponding module.

To enable style transfer using non-parallel training data, we enforce that
for a sample $\x_s \in \X_s$, its decoded sequence using $\mb{G}$ given its content representation $\z$ and the target style $\y^*$ should be in the target domain $\mathcal{X}_t$.
We use the idea of GAN~\citep{goodfellow2014generative}) and introduce an adversarial loss to be minimized in decoding.
The goal of the discriminator $\mb{D}$ is to distinguish between $\mb{G}(\mb{z}_s, \mb{y}^*)$ and $\mb{G}(\mb{z}_t, \mb{y}^*)$, while the generator tries to bewilder the discriminator:
\begin{align} \label{eq:adv}
\mathcal{L}_{adv} (&\bm{\theta}_{\mb{D}}, \bm{\theta}_{\mb{G}}, \bm{\theta}_{\mb{E}_{\z}}, \bm{\theta}_{\mb{E}_{\y}})\nonumber\\
=& \mathbb{E}_{\mb{x}_{s} \sim \X_s}\left[-\log (1 - \mb{D}(\mb{G}(\mb{z}_{s}, \mb{y}^*)))\right] \nonumber \\
+& \mathbb{E}_{\mb{x}_{t} \sim \mb{X}_{t}}\left[-\log \mb{D}(\mb{G}(\mb{z}_{t}, \mb{y}^*))\right] \,.
\end{align}
{\bf Remarks:} The above encoder-decoder framework is under-constrained in two aspects:
\begin{itemize}[wide=0\parindent,noitemsep]
\item[1.] For a sample $\x_s \in \X_s$, $\y_s$ can be an arbitrary value that minimizes the above losses in Equation~\ref{eq:rec} and ~\ref{eq:adv}, which may not necessarily capture the sentence style. 
This will affect the other decomposed part $\z$, making it not fully represent the content which should be invariant with the style.
\item[2.] The discriminator can only encourage the generated sentence to be aligned with the target domain $\mathcal{X}_2$, but cannot guarantee to keep the content of the source sentence intact. 
\end{itemize}

To address the first problem, we propose a style discrepancy loss, to constrain that the learnt $\y$ should have its distance from $\y^*$ guided by another discriminator
which evaluates the closeness of the sentence style to the target style.
For the second problem, we get inspired by the idea in \citet{he2016dual}  and \citet{zhu2017unpaired} and introduce a cycle consistency loss applicable to word sequence,
which requires that the generated sentence $\tilde{\mb{x}}$ can be transferred back to the original sentence $\mb{x}$. 

\subsection{Style Discrepancy Loss}

By using a portion of the training data, we can first train a discriminator $\mb{D}_s$ to predict whether a given sentence $\x$ has the target language style with
an output probability, denoted as $p_{\mb{D}_s}(\mb{x} \in \mathcal{X}_t)$.
When learning the decomposed style representation $\y_s$ for a sample  $\x_s \in \X_s$,
we enforce that the discrepancy between this style representation
and the target style representation $\y^*$, should be consistent with the output probability from $\mb{D}_s$. 
Specifically, since the styles are represented with embedding vectors, we measure the style discrepancy using the $\ell_2$ norm:
\begin{align} \label{eq:sd_learner}
d(\y_s,\y^*) = \|\mb{y}_s - \mb{y}^*\|_{2} \,. 
\end{align}
Intuitively, if a sentence has a larger probability to be considered having the target style, its style representation should be closer to the target style representation $\y^*$.
Thus, we would like to have $d(\y_s,\y^*)$ positively correlated with $1 - p_{\mb{D}_s}(\mb{x}_s \in \mathcal{X}_t)$. 
To incorporate this idea in our model, 
we use a probability density function $q(\mb{y}_s, \mb{y}^*)$, and define the style discrepancy loss as:
\begin{align}\label{eq:sd_loss}
\mathcal{L}_{dis} (&\bm{\theta}_{\mb{E}_{\y}}) \\
=& \mathbb{E}_{\mb{x}_{s} \sim \X_{s}}[-p_{\mb{D}_s}(\mb{x}_s \in \mathcal{X}_t) \log q(\mb{y}_s, \mb{y}^*)]\,, \nonumber
\end{align} 
where $q(\mb{y}_s, \mb{y}^*) = f(d (\y_s,\y^*))$ ($f(\cdot)$ is a valid probability density function) and
$\mb{D}_s$ is pre-trained and then fixed.
If a sentence $\mb{x}_s$ has a large $p_{\mb{D}_s}(\mb{x}_s \in \mathcal{X}_t)$, incorporating the above loss into the encoder-decoder framework will encourage a large $q(\mb{y}_s, \mb{y}^*)$ and hence a small $d(\y_s,\y^*)$, which means $\y_s$ will be close to $\y^*$.
In our experiment, we instantiate $q(\mb{y}_s, \mb{y}^*)$  with the standard normal distribution for simplicity:
\begin{align}\label{eq:sd_q}
q(\mb{y}_s, \mb{y}^*) = \frac{1}{\sqrt{2\pi}}  \exp(-\frac{d(\y_s,\y^*)^{2}}{2}).
\end{align}
However, better probability density functions can be used if we have some prior knowledge of the style distribution. With Equation~\ref{eq:sd_q}, the style discrepancy loss can be equivalently minimized by:
\begin{align}\label{eq:sd_equal_loss}
\mathcal{L}_{dis} (&\bm{\theta}_{\mb{E}_{\y}}) \\
=& \mathbb{E}_{\mb{x}_{s} \sim \X_{s}}[p_{\mb{D}_s}(\mb{x}_s \in \mathcal{X}_t) d(\mb{y}_s, \mb{y}^*)^{2}] \,.\nonumber
\end{align} 

Note that $\mb{D}_s$ is not jointly trained in our model. 
The reason is that, if we integrate it into the end-to-end training, we may start with a $\mb{D}_s$ with a low accuracy, and then our model is inclined to optimize a wrong style-discrepancy loss for many epochs and get stuck into a poor local optimum.

\subsection{Cycle Consistency Loss} \label{sec:cycle}


Inspired by \citet{he2016dual,zhu2017unpaired}, we require that a sentence transferred by the generator $\mb{G}$ should preserve the content of its original sentence, and thus it should have the capacity to recover the original sentence in a cyclic manner. 
For a sample $\x_s \in \X_s$ with its transferred sentence $\tilde{\x}_s$ having the target style $\y^*$, we encode $\tilde{\x}_s$ and combine its content $\tilde{\z}_s$ with its original style $\y_s$ for decoding.
We should expect that with a high probability, the original sentence $\x_1$ is generated.
For a sample $\x_t \in \X_t$, though we do not aim to change its language style in our task, we can still compute its cycle consistency loss for the purpose of additional regularization. We first choose an arbitrary style $\y_s$ obtained from a sentence in $\X_s$, and transfer $\x_t$ into this $\y_s$ style.
Next, we put this generated sentence into the encoder-decoder model with the style $\y^*$, and the original sentence $\x_t$ should be generated.
Formally, the cycle consistency loss is:
\begin{align}
\mathcal{L}_{cyc} (&\bm{\theta}_{\mb{E}_{\z}},\bm{\theta}_{\mb{E}_{\y}}, \bm{\theta}_{\mb{G}}) \nonumber\\
=&\mathbb{E}_{\mb{x}_{s} \sim \X_{s}} \left[-\log p_{\mb{G}}(\mb{x}_{s} | \mb{E}_{\z}(\tilde{\mb{x}}_s), \mb{y}_{s})\right] \nonumber \\
+&\mathbb{E}_{\mb{x}_{t} \sim \X_{t}} \left[-\log p_{\mb{G}}(\mb{x}_{t} | \mb{E}_{\z}(\tilde{\mb{x}}_{t}) ,\mb{y}^*)\right]\,.
\end{align}

\subsection{Full Objective}
An illustration of our basic model with the style discrepancy loss is shown in Figure~\ref{fig:model1} and the full model combined with the cycle consistency loss is shown in Figure~\ref{fig:model2}. To summarize, the full loss function of our model is:
\begin{align}
\mathcal{L} (&\bm{\theta}_{\mb{E}_{\z}}, \bm{\theta}_{\mb{E}_{\y}}, \bm{\theta}_{\mb{G}}, \bm{\theta}_{\mb{D}}) \nonumber \\
=& \mathcal{L}_{rec} - \lambda_1 \mathcal{L}_{adv} + \lambda_2 \mathcal{L}_{cyc} + \lambda_3 \mathcal{L}_{dis},
\end{align}
where $\lambda_{1}, \lambda_{2}, \lambda_{3}$ are parameters balancing the relative importance of the different loss parts. The overall training objective is a minmax game played among the encoder $\mb{E}_{\z}$, $\mb{E}_{\y}$, generator $\mb{G}$ and discriminator $\mb{D}$:
\begin{align}
\min_{\mb{E}_{\z}, \mb{E}_{\y}, \mb{G} } \max_{\mb{D}} \mathcal{L}(\bm{\theta}_{\mb{E}_{\z}}, \bm{\theta}_{\mb{E}_{\y}}, \bm{\theta}_{\mb{G}}, \bm{\theta}_{\mb{D}}) \,.
\end{align}
We implement the encoder $\mb{E}_{\z}$ using an RNN
with the last hidden state as the content representation,
and the style encoder $g(\x)$ using a CNN with the output representation of the last layer as the style representation.
The generator $\mb{G}$ is an RNN that takes the concatenation of the content and style representations as the initial hidden state.
The discriminator $\mb{D}$ and the pre-trained discriminator $\mb{D}_s$ used in the style discrepancy loss are CNNs with the similar network structure in $\mb{E}_{\y}$ followed by a sigmoid output layer.

\section{Experiments}


\subsection{Datasets}
\textbf{Yelp:}
Raw data are from the \textit{Yelp Dataset Challenge Round 10}, which are restaurant reviews on Yelp. Generally, reviews rated with 4 or 5 stars are considered positive, 1 or 2 stars are negative, and 3 stars are neutral.
For positive and negative reviews, we use the processed data released by~\citet{shen2017style}, which contains 250k negative sentences and 350k positive sentences.
For neutral reviews, we follow similar steps in~\citet{shen2017style} to process and select the data.
We first filter out neutral reviews (rated with 3 stars and categorized with the keyword `restaurant') with the length exceeding 15 or less than 3. Then, data selection in~\citet{moore2010intelligent} is used to ensure a large enough vocabulary overlap between the neutral data and the data in~\citet{shen2017style}. Afterwards, we sample 500k sentences from the resulting dataset as the neutral data.
We use the positive data as the target style domain.
Based on the three classes of data, we construct two source datasets with multiple styles:
\begin{itemize}[wide=0\parindent,noitemsep]
	\item Positive+Negative (Pos+Neg):
	we add different numbers of positive data (50k, 100k, 150k) into the negative data
	so that the source domain contains data with two sentiments.
	\item Neutral+Negative (Neu+Neg):
	we combine neutral (50k, 100k, 150k) and negative data together as the source data.
\end{itemize}
We consider the Neu+Neg dataset is harder to learn from than the Pos+Neg dataset since for the Pos+Neg dataset, we can make use of a pre-trained classifier to possibly filter out some positive data so that most of the source data have the same style and the model in \citet{shen2017style} can work. However, the neutral data cannot be removed in this way. Also,
most of the real data may be in the neutral sentiment, and we want to see if such sentences can be transferred well.

\noindent\textbf{Chat:} 
We use sentences from a real Chinese dialog dataset as the source  domain.
Users can chat with various personalized language styles, which are not easy to be classified
into one of the three sentiments as in Yelp.
Romantic sentences are collected from several online novel websites and filtered by human annotators. 
The dataset has 400k romantic sentences and 800k general sentences.
Our task is to transfer the dialog sentences to a romantic style, characterized by the selected romantic sentences.

\noindent\textbf{Shakespeare:} 
We experiment on revising modern text in the Shakespearean style at the sentence-level as in~\cite{mueller2017sequence}. Following their experimental setup, we collect 29,388 sentences authored by Shakespeare and 54,800 sentences from non-Shakespeare-authored works. The length of all the sentences ranges from 3 to 15.

Each of the above three datasets is divided into three non-overlapping parts which are respectively used for the style transfer model, the pre-trained discriminator $\mb{D}_{s}$, and the evaluation classifier used in Section~\ref{sec:eval_setup}. Each part is further divided into training, testing, and validation sets. There is one exception that $\mb{D}_{s}$ for Shakespeare is trained on a subset of the data for training the style transfer model because the Shakespeare dataset is small.
Statistics of the data for training and evaluating the style transfer model are shown in the supplementary material.

\subsection{Compared Methods and Configurations}

We compare our method with \citet{shen2017style}
which is the state-of-the-art language style transfer model with non-parallel data, and we name as Style Transfer Baseline (STB).
As described in Section~\ref{sec:related} and Section~\ref{sec:model}, STB is built upon an auto-encoder framework. 
It focuses on transferring sentences from one style to the other, with 
the source and target language styles represented by two embedding vectors. 
It also relies on adversarial training methods to align content spaces of the two domains.
We keep the configurations of the modules in STB, such as the encoder, decoder and discriminator, the same as ours for a fair comparison.


We implement our model using Tensorflow~\citep{abadi2016tensorflow}.
We use GRU as the encoder and generation cells in our encoder-decoder framework.
Dropout~\citep{srivastava2014dropout} is applied in GRUs and the dropout probability is set to 0.5. 
Throughout our experiments, we set the dimension of the word embedding, content representation and style representation as $200$, $1000$ and $500$ respectively.
For the style encoder $g(\x)$, we follow the CNN architecture in \citet{kim2014convolutional}, 
and use filter sizes of $200 \times \{1, 2, 3, 4, 5\}$ with 100 feature maps each, so that the resulting output layer is of size $500$, i.e., the dimension of the style representation. 
The pre-trained discriminator $\mb{D}_{s}$ is implemented similar to $g(\x)$ but using filter sizes $200 \times \{2, 3, 4, 5\}$ with 250 feature maps each.
The testing accuracy of the pre-trained $\mb{D}_{s}$ is 95.23\% for Yelp, 87.60\% for Chat, and 87.60\% for Shakespeare.
We further set the balancing parameters $\lambda_{1}=\lambda_{2}=1$, $\lambda_{3}=5$, and train the model using the Adam optimizer~\citep{kingma2014adam} with the learning rate $10^{-4}$.
All input sentences are padded so that they have the same length 20 for Yelp and Shakespeare and 35 for Chat. 
Furthermore, we use the pre-trained word embeddings \texttt{Glove}~\citep{pennington2014glove} for Yelp and Shakespeare and use the Chinese word embeddings trained on a large amount of Chinese news data for Chat when training the classifiers.

\subsection{Evaluation Setup }\label{sec:eval_setup}
Following \citet{shen2017style}, we use a model-based evaluation metric. Specifically, we use a pre-trained evaluation classifier to classify whether the transferred sentence has the correct style.
The evaluation classifier is implemented with the same network structure as the discriminator $\mb{D}_{s}$.
The testing accuracy of evaluation classifiers is 95.36\% for Yelp, 87.05\% for Chat, and 88.70\% for Shakespeare. 
We repeat the training three times for each experiment setting and report the mean accuracy on the testing data with their standard deviation.
 

\subsection{Experiment Roadmap}
We perform a set of experiments on Yelp, which is also used in \citet{shen2017style}, to systematically investigate the effectiveness of our model:
\begin{itemize}[wide=0\parindent,noitemsep]
\item[1.]We validate the usefulness of the proposed style discrepancy loss and the cycle consistency loss in our setting (Sec~\ref{sec:exp_loss}).
\item[2.]We vary the difficulty level of the source data and verify the robustness of our model (Sec~\ref{sec:exp_diff}).
\item[3.]We look into some transferred sentences and analyze successful cases and failed cases (Sec~\ref{sec:exp_case}).
\item[4.]We perform human evaluations to evaluate the overall quality of the transferred sentences and check if the results are consistent with those using the model-based evaluation (Sec~\ref{sec:exp_man}).
\end{itemize}
After conducting the thorough study of our model on Yelp,
we apply our model on Chat and Shakespeare, two real datasets in which the source domain has various language styles.

\subsection{Experiments and Analysis on Yelp}
\subsubsection{Ablation Study}\label{sec:exp_loss}
To investigate the effectiveness of the style discrepancy loss, we try our full model by removing the style discrepancy loss only on the dataset Pos+Neg whose source domain contains both positive sentences and negative sentences.
When the model converges, we find that the cycle consistency loss is much larger than that obtained in the full model for the same setting. 
We also manually check that almost all sentences fail to transfer their styles with this model. 
This indicates that the proposed style discrepancy loss is an inseparable part of our model. 
\begin{table}[!ht]\small
	\begin{center}
		{\setlength{\tabcolsep}{.4em}
			\makebox[\linewidth]{\resizebox{\linewidth}{!}{%
					\begin{tabular}{ccccc}
						\toprule
						\#positive &  & STB & Ours &  \\
						samples used & STB & (with Cyc) & (without Cyc) & Ours \\
						\midrule
						50k & $0.904\pm0.033$ & $0.912\pm0.017$ & $0.890\pm0.027$ &  $\mb{0.943\pm0.007}$ \\
						100k & $0.800\pm0.043$ & $0.846\pm0.031$ & $0.862\pm0.010$ & $\mb{0.940\pm0.005}$ \\
						150k & $0.535\pm0.086$ & $0.678\pm0.033$ & $0.631\pm0.168$ & $\mb{0.934\pm0.005}$ \\
						\bottomrule
		\end{tabular}}}}
		\vskip -.05in
	\caption{\label{tab:yelp_pos}Testing accuracies on Yelp with Pos+Neg source data.}
	\end{center}
	\vskip -.12in
\end{table}		

We also validate the effectiveness of the cycle consistency loss
\footnote{Note that our proposed cycle consistency loss can be similarly added in STB.}.
Specifically, we compare two versions of both STB and our model, one with the cycle consistency loss and one without.
 We vary the number of positive sentences in the source domain 
and results are shown in Table~\ref{tab:yelp_pos}. 
It can be seen that incorporating the cycle consistency loss consistently improves the performance for both STB and our proposed model.

\subsubsection{Difficulty Level of Training Data}\label{sec:exp_diff}

\noindent\textbf{Pos+Neg as Source Data:}We compare STB and our proposed model on the first dataset Pos+Neg using the results in Table~\ref{tab:yelp_pos}. 
As the number of positive sentences in the source data increases, the average performance of both versions of STB decreases drastically.
This is reasonable because STB introduces a discriminator to align the sentences from the target domain back to the source domain, and when the source domain contains more positive samples, it is hard to find a good alignment to the source domain.
Meanwhile the performance of our model, even the basic one without the cycle consistency loss, does not fluctuate much with the increase of the number of positive samples, showing that our model is not that sensitive to the source data containing more than one sentiments. 
Overall, our model with the cycle consistency loss performs the best.

\begin{table}[!ht]\small
	\begin{center}
				{\setlength{\tabcolsep}{.4em}
			\makebox[\linewidth]{\resizebox{\linewidth}{!}{%
		\begin{tabular}{cp{.2\textwidth}<{\centering}c}
			\toprule
			\#Neural samples & STB (with Cyc)  & Ours \\
			\midrule
			50k   & $0.929\pm0.011$   & $\mb{0.946\pm0.001}$ \\
			100k   & $0.937\pm0.018$  & $\mb{0.939\pm0.000}$ \\
			150k   & $0.926\pm0.019$  & $\mb{0.936\pm0.006}$ \\
			\bottomrule
		\end{tabular}}}}
	\vskip -.05in
	\caption{\label{tab:yelp_neu}Testing accuracies on Yelp with Neu+Neg source data.}
	\end{center}
	\vskip -.12in
\end{table}	

\noindent\textbf{Neu+Neg as Source Data:} The dataset Pos+Neg is not so challenging because we can use a pre-trained discriminator similar to $\mb{D}_s$ in our model, to remove those samples classified as positive with high probabilities, so that only sentences with a less positive sentiment remain in the source domain.
Thus, we test on our second dataset Neu+Neg.
In this setting, in case that some positive sentences exist in those neutral reviews, when STB is trained, we use the same pre-trained discriminator in our model to filter out samples classified as positive with probabilities larger than 0.9.
In comparison, our model uses all data, since it naturally allows for those data with styles similar to the target style.
Experimental results in Table~\ref{tab:yelp_neu} show that
with the same amount of neutral data mixed in the source domain, our model performs better than STB (with Cyc) and is relatively stable among multiple runs.

\noindent\textbf{Limited Sentences in Target Domain:} In real applications, there may be only a small amount of data in the target domain. To simulate this scenario, we limit the amount of the target data (randomly sampled from the positive data) used for training, and evaluate the robustness of the compared methods. 
Table~\ref{tab:yelp_target} shows the experimental results. 
It is surprising to see that both methods obtain relatively steady accuracies with different numbers of target samples.
Yet, our model surpasses STB (with Cyc) in all the cases. 
\begin{table}[!ht]\small
	\begin{center}
				{\setlength{\tabcolsep}{.4em}
			\makebox[\linewidth]{\resizebox{\linewidth}{!}{%
		\begin{tabular}{ccc}
			\toprule
			\#Target samples used&STB (with Cyc) &Ours \\
			\midrule
			100k  & $0.742\pm0.009$  & $\mb{0.894\pm0.010}$ \\
			150k  & $0.756\pm0.020$  & $\mb{0.907\pm0.003}$ \\
			200k  & $0.735\pm0.022$  & $\mb{0.906\pm0.012}$ \\
			\bottomrule
		\end{tabular}}}}
	\caption{\label{tab:yelp_target}Testing accuracies on Yelp with different numbers of target samples used.}
	\end{center}
\end{table}	

\subsubsection{Case Study}\label{sec:exp_case}

We manually examine the generated sentences for a detailed study.
Overall, our full model can generate grammatically correct positive reviews without changing the original content in more cases than the other methods.
In Table~\ref{tab:yelp_example}, we present some example results of the various methods. 
We can see that when the original sentences are simple such as the first example,
all models can transfer the sentence successfully.
However, when the original sentences are complex,
both versions of STB and our basic model (without the cycle consistency loss) cannot generate fluent sentences, 
but our full model still succeeds.
One minor problem with our model is that it may use a wrong tense in transferred sentences (i.e., the last transferred sentence by our model),
which however does not influence the sentence style and meaning a lot.

\begin{table}[!ht]\small
\begin{center}
{\setlength{\tabcolsep}{.4em}
\makebox[\linewidth]{\resizebox{\linewidth}{!}{%
\begin{tabular}{cl}
	\toprule
	Original Sentence & and just not very good . \\
	\midrule
	STB & but i was very good . \\
	STB (with Cyc) & and just always very good . \\
	Ours (without Cyc)& and just always very good .  \\
	Ours & and i love it . \\
	\toprule
	Original Sentence & i have tried to go to them twice silly me . \\
	\midrule
	STB & i 'm going to anyone when they need to .  \\
	STB (with Cyc) & i 've been to go here for years out . \\
	Ours (without Cyc)& i have recommend to anyone 's your home needs .  \\
	Ours & i have tried the place and it was great . \\
	\toprule
	Original Sentence & i am so thankful to be out of this place . \\
	\midrule
	STB &  i am so impressed to the experience i had .\\
	STB (with Cyc) & i am always impressed to see for a family .\\
	Ours (without Cyc) & i am so grateful for being on this place again .\\
	Ours & i am so happy to have found this place .\\
	\toprule
	Original Sentence & service was okay but joseph is just rude . \\
	\midrule
	STB & service was well , but well . \\
	STB (with Cyc) & service was friendly and everyone is .\\
	Ours (without Cyc) & service was quick , and just funny .\\
	Ours & service is great and food was great .\\
	\toprule
	Original Sentence & they were very loud and made noise throughout the night . \\
	\midrule
	STB &  they were very good and made my own water .\\
	STB (with Cyc) & they were very good and made out on the game .\\
	Ours (without Cyc) & they were very nice and made the other time .\\
	Ours & they are very friendly and made you feel comfortable .\\
	\bottomrule
\end{tabular}}}}
\caption[LOF]{\label{tab:yelp_example}Example sentences on Yelp transferred into a positive sentiment.}
\end{center}
\end{table}	

\subsubsection{Human Evaluation}\label{sec:exp_man}
The model-based evaluation metric is inadequate at measuring whether a transferred sentence preserves the content of a source sentence~\citep{fu2017style}.
Therefore, we rely on human evaluations to evaluate model performance in content preservation.
We randomly select 200 test samples from Yelp and perform human evaluations to estimate the overall quality of transferred sentences rating from 1 (failed), 2 (tolerable), 3 (satisfying) and 4 (perfect) by jointly considering content preservation, sentiment modification, and fluency of the transferred sentences. 


We hire five annotators to evaluate the results. Since we have 9 settings and totally 24 different methods in Table~\ref{tab:yelp_pos}-\ref{tab:yelp_target}, 
here we select one of the settings on Yelp due to limited budgets.
Table~\ref{tab:eval_shit} 
shows the averaged scores of the annotator's evaluations. 
As can be seen, 
by considering all the above three aspects, 
our model is better than other methods.
This result is also consistent with the automatic evaluation ones.

\begin{table}[!ht]
	\begin{center}		
		{\setlength{\tabcolsep}{.4em}
			\makebox[\linewidth]{\resizebox{\linewidth}{!}{%
					\begin{tabular}{ccccc}
						\toprule
						& STB & (with Cyc) & (without Cyc) & Ours \\
						\midrule
						Overall & $2.352\pm0.288$ & $2.545\pm0.206$  & $2.771\pm0.290$ &  $\mb{2.805\pm0.314}$ \\
						\bottomrule
		\end{tabular}}}}
	\caption{\label{tab:eval_shit}Human evaluation on Yelp when 150k positive sentences are added to source domain  (row 3 in Table~\ref{tab:yelp_pos}).}
	\end{center}
\end{table}		

\begin{table}[!ht]
	\begin{center}
		{\setlength{\tabcolsep}{.4em}
			\makebox[\linewidth]{\resizebox{\linewidth}{!}{%
					\begin{tabular}{ccc}
						\toprule
						\#target samples used & STB (with Cyc) & Ours \\
						\midrule
						10k & $0.880\pm0.016$ & $\mb{0.962\pm0.002}$ \\
						50k & $0.927\pm0.012$ & $\mb{0.980\pm0.003}$ \\
						100k & $0.940\pm0.005$ & $\mb{0.968\pm0.002}$ \\
						150k & $0.943\pm0.004$ & $\mb{0.965\pm0.003}$ \\
						\bottomrule
		\end{tabular}}}}
		\vskip -.05in
		\caption{\label{tab:chat_target}Testing accuracies on Chat with different numbers of target samples used.}
	\end{center}
\end{table}

\subsection{Experiments and Analysis on Chat}
As in the Yelp experiment, we vary the number of target sentences to test the robustness of the compared methods.
The experimental results are shown in Table~\ref{tab:chat_target}. 
Several observations can be made. First, STB (with Cyc) obtains a relatively low performance with only 10k target samples. As more target samples are used, its performance increases. 
Second, our model achieves a high accuracy even with 10k target samples used, and remains stable in all the cases.
Thus, our model achieves better performance as well as stronger robustness on Chat.
Due to space limitations, we present a few examples in Table 10 in the supplementary material.
We find that our model generally successfully transfers the sentence into a romantic style with some romantic phrases used.

\subsection{Experiments and Analysis on Shakespeare}

The testing accuracies are shown in Table~\ref{tab:sksp_target}.
We also present some example sentences in Table~\ref{tab:sksp_example}.
Compared with STB, our model can generate sentences which are more fluent and have a higher probability to have a correct target style. 
For example, in the sentences transferred by our model, the words such as \textit{'sir'}, \textit{'master'}, and \textit{'lustre'} are used, which are common in the Shakespearean works.
However, we find that both STB and our model tend to generate short sentences and change the content of source sentences in more cases in this set of experiment than in the Yelp and Chat datasets. 
We conjecture this is caused by the scarcity of training data. Sentences in the Shakespearean style form a vocabulary of 8559 words, but almost 60\% of them appear less than 10 times. 
On the other hand, the source domain contains 19962 words, but there are only 5211 common words in these two vocabularies. Thus aligned words/phrases may not exist in the dataset.

\begin{table}[!ht]
\begin{center}

{\setlength{\tabcolsep}{.4em}
\makebox[\linewidth]{\resizebox{\linewidth}{!}{%
		\begin{tabular}{ccc}
			\toprule
			\#target samples used & STB (with Cyc) & Ours \\
			\midrule
			21,888 & $0.956\pm0.014$ & $0.965\pm0.009$ \\
			\bottomrule
\end{tabular}}}}
\caption{\label{tab:sksp_target}Testing accuracies on Shakespeare.}
\end{center}
\end{table}

\begin{table}[!ht]
\vspace{-1em}
\begin{center}

\makebox[\linewidth]{\resizebox{\linewidth}{!}{%
		\begin{tabular}{p{0.2\textwidth}<{\centering}lp{.0\textwidth}<{\centering}}
			\toprule
			Original Sentence & i should never have thought of such a thing . & \\
			\midrule
			STB (with Cyc) & i shall not have to for you . & \\
			Ours & i will never be thee , sir . & \\
			\toprule
			Original Sentence & do n't try to make any stupid moves . & \\
			\midrule
			STB (with Cyc) & i think you have to your . & \\
			Ours & do you walk , the master ? & \\
			\toprule
			Original Sentence & where were the hardships she had expected ? & \\
			\midrule
			STB (with Cyc) & what is it is it ? & \\
			Ours & where are the lustre here ? & \\
			\bottomrule
\end{tabular}}}
\caption{\label{tab:sksp_example}Example non-Shakespeare sentences transferred into a Shakespearean language style.}
\end{center}
\vskip -.12in
\end{table}	

\section{Conclusion}

We have presented an encoder-decoder framework for language style transfer. It allows for the use of non-parallel data, where the source data have various unknown language styles. 
Each sentence is encoded into two latent representations, one corresponding to its content disentangled from the style and
and the other representing the style only.
By recombining the content with the target style, we can decode a sentence aligned in the target domain.
Specifically, we propose two loss functions, i.e., the style discrepancy loss and the cycle consistency loss, to adequately constrain the encoding and decoding functions.
The style discrepancy loss is used to enforce a properly encoded style representation while the cycle consistency loss is utilized to ensure that the style-transferred sentences can be transferred back to their original sentences.
Experimental results in three tasks demonstrate that our proposed method outperforms previous style transfer methods.


\bibliography{emnlp2018}

\begin{thebibliography}{28}
\expandafter\ifx\csname natexlab\endcsname\relax\def\natexlab#1{#1}\fi

\bibitem[{Abadi et~al.(2016)Abadi, Agarwal, Barham, Brevdo, Chen, Citro,
  Corrado, Davis, Dean, Devin et~al.}]{abadi2016tensorflow}
Mart{\'\i}n Abadi, Ashish Agarwal, Paul Barham, Eugene Brevdo, Zhifeng Chen,
  Craig Citro, Greg~S Corrado, Andy Davis, Jeffrey Dean, Matthieu Devin, et~al.
  2016.
\newblock Tensorflow: Large-scale machine learning on heterogeneous distributed
  systems.
\newblock \emph{arXiv preprint arXiv:1603.04467}.

\bibitem[{Bahdanau et~al.(2014)Bahdanau, Cho, and Bengio}]{bahdanau2014neural}
Dzmitry Bahdanau, Kyunghyun Cho, and Yoshua Bengio. 2014.
\newblock Neural machine translation by jointly learning to align and
  translate.
\newblock \emph{arXiv preprint arXiv:1409.0473}.

\bibitem[{Chung et~al.(2014)Chung, Gulcehre, Cho, and
  Bengio}]{chung2014empirical}
Junyoung Chung, Caglar Gulcehre, KyungHyun Cho, and Yoshua Bengio. 2014.
\newblock Empirical evaluation of gated recurrent neural networks on sequence
  modeling.
\newblock \emph{arXiv preprint arXiv:1412.3555}.

\bibitem[{Ficler and Goldberg(2017)}]{ficler2017controlling}
Jessica Ficler and Yoav Goldberg. 2017.
\newblock Controlling linguistic style aspects in neural language generation.
\newblock In \emph{Proceedings of the Workshop on Stylistic Variation}, pages
  94--104.

\bibitem[{Fu et~al.(2017)Fu, Tan, Peng, Zhao, and Yan}]{fu2017style}
Zhenxin Fu, Xiaoye Tan, Nanyun Peng, Dongyan Zhao, and Rui Yan. 2017.
\newblock Style transfer in text: Exploration and evaluation.
\newblock \emph{arXiv preprint arXiv:1711.06861}.

\bibitem[{Gatys et~al.(2016)Gatys, Ecker, and Bethge}]{gatys2016image}
Leon~A Gatys, Alexander~S Ecker, and Matthias Bethge. 2016.
\newblock Image style transfer using convolutional neural networks.
\newblock In \emph{Proceedings of the IEEE Conference on Computer Vision and
  Pattern Recognition}, pages 2414--2423.

\bibitem[{Goodfellow et~al.(2014)Goodfellow, Pouget-Abadie, Mirza, Xu,
  Warde-Farley, Ozair, Courville, and Bengio}]{goodfellow2014generative}
Ian Goodfellow, Jean Pouget-Abadie, Mehdi Mirza, Bing Xu, David Warde-Farley,
  Sherjil Ozair, Aaron Courville, and Yoshua Bengio. 2014.
\newblock Generative adversarial nets.
\newblock In \emph{Advances in Neural Information Processing Systems}, pages
  2672--2680.

\bibitem[{He et~al.(2016)He, Xia, Qin, Wang, Yu, Liu, and Ma}]{he2016dual}
Di~He, Yingce Xia, Tao Qin, Liwei Wang, Nenghai Yu, Tieyan Liu, and Wei-Ying
  Ma. 2016.
\newblock Dual learning for machine translation.
\newblock In \emph{Advances in Neural Information Processing Systems}, pages
  820--828.

\bibitem[{Hu et~al.(2017)Hu, Yang, Liang, Salakhutdinov, and
  Xing}]{hu2017toward}
Zhiting Hu, Zichao Yang, Xiaodan Liang, Ruslan Salakhutdinov, and Eric~P Xing.
  2017.
\newblock Toward controlled generation of text.
\newblock In \emph{Proceedings of the International Conference on Machine
  Learning}, pages 1587--1596.

\bibitem[{Kim(2014)}]{kim2014convolutional}
Yoon Kim. 2014.
\newblock Convolutional neural networks for sentence classification.
\newblock In \emph{Proceedings of the Conference on Empirical Methods in
  Natural Language Processing}, pages 1746--1751.

\bibitem[{Kingma and Ba(2015)}]{kingma2014adam}
Diederik Kingma and Jimmy Ba. 2015.
\newblock Adam: A method for stochastic optimization.
\newblock In \emph{Proceedings of the International Conference for Learning
  Representations}.

\bibitem[{Kingma and Welling(2013)}]{kingma2013auto}
Diederik~P Kingma and Max Welling. 2013.
\newblock Auto-encoding variational bayes.
\newblock \emph{arXiv preprint arXiv:1312.6114}.

\bibitem[{Kulkarni et~al.(2015)Kulkarni, Whitney, Kohli, and
  Tenenbaum}]{kulkarni2015deep}
Tejas~D Kulkarni, William~F Whitney, Pushmeet Kohli, and Josh Tenenbaum. 2015.
\newblock Deep convolutional inverse graphics network.
\newblock In \emph{Advances in Neural Information Processing Systems}, pages
  2539--2547.

\bibitem[{Li et~al.(2016)Li, Galley, Brockett, Spithourakis, Gao, and
  Dolan}]{li2016persona}
Jiwei Li, Michel Galley, Chris Brockett, Georgios~P Spithourakis, Jianfeng Gao,
  and Bill Dolan. 2016.
\newblock A persona-based neural conversation model.
\newblock In \emph{Proceedings of the Annual Meeting of the Association for
  Computational Linguistics}, pages 994--1003.

\bibitem[{Li et~al.(2018)Li, Jia, He, and Liang}]{li2018delete}
Juncen Li, Robin Jia, He~He, and Percy Liang. 2018.
\newblock Delete, retrieve, generate: A simple approach to sentiment and style
  transfer.
\newblock \emph{arXiv preprint arXiv:1804.06437}.

\bibitem[{Liu and Tuzel(2016)}]{liu2016coupled}
Ming-Yu Liu and Oncel Tuzel. 2016.
\newblock Coupled generative adversarial networks.
\newblock In \emph{Advances in Neural Information Processing Systems}, pages
  469--477.

\bibitem[{Luan et~al.(2017)Luan, Paris, Shechtman, and Bala}]{luan2017deep}
Fujun Luan, Sylvain Paris, Eli Shechtman, and Kavita Bala. 2017.
\newblock Deep photo style transfer.
\newblock \emph{arXiv preprint arXiv:1703.07511}.

\bibitem[{Moore and Lewis(2010)}]{moore2010intelligent}
Robert~C Moore and William Lewis. 2010.
\newblock Intelligent selection of language model training data.
\newblock In \emph{Proceedings of the Annual Meeting of the Association for
  Computational Linguistics}, pages 220--224.

\bibitem[{Mueller et~al.(2017)Mueller, Gifford, and
  Jaakkola}]{mueller2017sequence}
Jonas Mueller, David Gifford, and Tommi Jaakkola. 2017.
\newblock Sequence to better sequence: continuous revision of combinatorial
  structures.
\newblock In \emph{Proceedings of the International Conference on Machine
  Learning}, pages 2536--2544.

\bibitem[{Nallapati et~al.(2016)Nallapati, Zhou, Gulcehre, Xiang
  et~al.}]{nallapati2016abstractive}
Ramesh Nallapati, Bowen Zhou, Caglar Gulcehre, Bing Xiang, et~al. 2016.
\newblock Abstractive text summarization using sequence-to-sequence rnns and
  beyond.
\newblock \emph{arXiv preprint arXiv:1602.06023}.

\bibitem[{Paulus et~al.(2017)Paulus, Xiong, and Socher}]{paulus2017deep}
Romain Paulus, Caiming Xiong, and Richard Socher. 2017.
\newblock A deep reinforced model for abstractive summarization.
\newblock \emph{arXiv preprint arXiv:1705.04304}.

\bibitem[{Pennington et~al.(2014)Pennington, Socher, and
  Manning}]{pennington2014glove}
Jeffrey Pennington, Richard Socher, and Christopher Manning. 2014.
\newblock Glove: Global vectors for word representation.
\newblock In \emph{Proceedings of the Conference on Empirical Methods in
  Natural Language Processing}, pages 1532--1543.

\bibitem[{Rush et~al.(2015)Rush, Chopra, and Weston}]{rush2015neural}
Alexander~M Rush, Sumit Chopra, and Jason Weston. 2015.
\newblock A neural attention model for abstractive sentence summarization.
\newblock In \emph{Proceedings of the Conference on Empirical Methods in
  Natural Language Processing}, pages 379--389.

\bibitem[{Shen et~al.(2017)Shen, Lei, Barzilay, and Jaakkola}]{shen2017style}
Tianxiao Shen, Tao Lei, Regina Barzilay, and Tommi Jaakkola. 2017.
\newblock Style transfer from non-parallel text by cross-alignment.
\newblock In \emph{Advances in Neural Information Processing Systems}.

\bibitem[{Shum et~al.(2018)Shum, He, and Li}]{harry2018}
Heung{-}Yeung Shum, Xiaodong He, and Di~Li. 2018.
\newblock From eliza to xiaoice: Challenges and opportunities with social
  chatbots.
\newblock \emph{CoRR}.

\bibitem[{Srivastava et~al.(2014)Srivastava, Hinton, Krizhevsky, Sutskever, and
  Salakhutdinov}]{srivastava2014dropout}
Nitish Srivastava, Geoffrey~E Hinton, Alex Krizhevsky, Ilya Sutskever, and
  Ruslan Salakhutdinov. 2014.
\newblock Dropout: a simple way to prevent neural networks from overfitting.
\newblock \emph{Journal of machine learning research}, 15(1):1929--1958.

\bibitem[{Sutskever et~al.(2014)Sutskever, Vinyals, and
  Le}]{sutskever2014sequence}
Ilya Sutskever, Oriol Vinyals, and Quoc~V Le. 2014.
\newblock Sequence to sequence learning with neural networks.
\newblock In \emph{Advances in Neural Information Processing Systems}, pages
  3104--3112.

\bibitem[{Zhu et~al.(2017)Zhu, Park, Isola, and Efros}]{zhu2017unpaired}
Jun-Yan Zhu, Taesung Park, Phillip Isola, and Alexei~A Efros. 2017.
\newblock Unpaired image-to-image translation using cycle-consistent
  adversarial networks.
\newblock In \emph{Proceedings of the International Conference on Computer
  Vision}.

\end{thebibliography}
\bibliographystyle{acl_natbib_nourl}

\newpage
\clearpage
\appendix

{\section*{\centering Supplementary Materials}}
This supplementary material contains the following contents.
(1) Table~\ref{tab:yelp_st}--\ref{tab:sksp_evaluation} show statistics of the data used in the experiments. 
(2) Table~\ref{tab:chat_example} shows some transferred sentences from the testing data of Chat.

\begin{table}[!ht]
	\vspace{-0em}
	\begin{center}

		\makebox[\linewidth]{\resizebox{\linewidth}{!}{%
				\begin{tabular}{cp{.1\textwidth}<{\centering}p{.1\textwidth}<{\centering}p{.1\textwidth}<{\centering}}
					\toprule
					& Training & Test & Validation \\
					\midrule
					Positive & 240417 & 40000 & 20000   \\
					Negative & 151026 & 40000 & 20000  \\
					\bottomrule
		\end{tabular}}}
		\caption{\label{tab:yelp_st}Statistics of Yelp for the style transfer model}
	\end{center}
\end{table}	

\begin{table}[!ht]
	\begin{center}

		\makebox[\linewidth]{\resizebox{\linewidth}{!}{%
				\begin{tabular}{cp{.1\textwidth}<{\centering}p{.1\textwidth}<{\centering}p{.1\textwidth}<{\centering}}
					\toprule
					& Training & Test & Validation \\
					\midrule
					Romantic & 207312 & 40000 & 40000 \\
					General & 514460 & 40000 & 40000 \\
					\bottomrule
		\end{tabular}}}
		\caption{\label{tab:chat_st}Statistics of Chat for the style-transfer model}
	\end{center}
\end{table}

\begin{table}[!ht]
	\vspace{-0em}
	\begin{center}

		\makebox[\linewidth]{\resizebox{\linewidth}{!}{%
				\begin{tabular}{cp{.1\textwidth}<{\centering}p{.1\textwidth}<{\centering}p{.1\textwidth}<{\centering}}
					\toprule
					& Training & Test & Validation \\
					\midrule
					Shakespeare & 21888 & 1000 & 2000   \\
					Non-Shakespeare & 43800 & 1000 & 2000  \\
					\bottomrule
		\end{tabular}}}
		\caption{\label{tab:sksp_st}Statistics of Shakespeare for the style transfer model}
	\end{center}
\end{table}

\begin{table}[!ht]
	\vspace{-0em}
	\begin{center}

		\makebox[\linewidth]{\resizebox{\linewidth}{!}{%
				\begin{tabular}{cp{.1\textwidth}<{\centering}p{.1\textwidth}<{\centering}p{.1\textwidth}<{\centering}}
					\toprule
					& Training & Test & Validation \\
					\midrule
					Positive & 75000 & 5000 & 2500 \\
					Negative & 37500 & 5000 & 2500 \\
					\bottomrule
		\end{tabular}}}
		\caption{\label{tab:yelp_cls}Statistics of Yelp for the discriminator $\mb{D}_s$}
	\end{center}
\end{table}	

\begin{table}[!ht]
	\begin{center}

		\makebox[\linewidth]{\resizebox{\linewidth}{!}{%
				\begin{tabular}{cp{.1\textwidth}<{\centering}p{.1\textwidth}<{\centering}p{.1\textwidth}<{\centering}}
					\toprule
					& Training & Test & Validation \\
					\midrule
					Positive & 37500 & 1500 & 2250 \\
					Negative & 18750 & 1500 & 2250 \\
					\bottomrule
		\end{tabular}}}
		\caption{\label{tab:yelp_evaluation}Statistics of Yelp for the evaluation classifier}
	\end{center}
\end{table}	

\begin{table}[!ht]
	\vspace{-0em}
	\begin{center}

		\makebox[\linewidth]{\resizebox{\linewidth}{!}{%
				\begin{tabular}{cp{.1\textwidth}<{\centering}p{.1\textwidth}<{\centering}p{.1\textwidth}<{\centering}}
					\toprule
					& Training & Test & Validation \\
					\midrule
					Romantic & 100000 & 10000 & 10000 \\
					General & 200000 & 10000 & 10000 \\
					\bottomrule
		\end{tabular}}}
		\caption{\label{tab:chat_cls}Statistics of Chat for the discriminator $\mb{D}_s$}
	\end{center}
\end{table}	

\begin{table}[!ht] 
	\begin{center}

		\makebox[\linewidth]{\resizebox{\linewidth}{!}{%
				\begin{tabular}{cp{.1\textwidth}<{\centering}p{.1\textwidth}<{\centering}p{.1\textwidth}<{\centering}}
					\toprule
					& Training & Test & Validation \\
					\midrule
					Romantic & 50000 & 5000 & 5000 \\
					General & 100000 & 5000 & 5000 \\
					\bottomrule
		\end{tabular}}}
		\caption{\label{tab:chat_evaluation}Statistics of Chat for the evaluation classifier}
		
	\end{center}
\end{table}	

\begin{table}[!ht]
	\vspace{-0em}
	\begin{center}

		\makebox[\linewidth]{\resizebox{\linewidth}{!}{%
				\begin{tabular}{cp{.1\textwidth}<{\centering}p{.1\textwidth}<{\centering}p{.1\textwidth}<{\centering}}
					\toprule
					& Training & Test & Validation \\
					\midrule
					Shakespeare & 3500 & 500 & 500 \\
					Non-Shakespeare & 7000 & 500 & 500 \\
					\bottomrule
		\end{tabular}}}
		\caption{\label{tab:sksp_cls}Statistics of Shakespeare for the discriminator $\mb{D}_s$}
	\end{center}
\end{table}	

\begin{table}[!ht] 
	\begin{center}

		\makebox[\linewidth]{\resizebox{\linewidth}{!}{%
				\begin{tabular}{cp{.1\textwidth}<{\centering}p{.1\textwidth}<{\centering}p{.1\textwidth}<{\centering}}
					\toprule
					& Training & Test & Validation \\
					\midrule
					Shakespeare & 3500 & 500 & 500 \\
					Non-Shakespeare & 7000 & 500 & 500 \\
					\bottomrule
		\end{tabular}}}
		\caption{\label{tab:sksp_evaluation}Statistics of Shakespeare for the evaluation classifier}
		
	\end{center}
\end{table}

\begin{CJK*}{UTF8}{gbsn}
	\begin{table*}[!ht]
		\begin{center}

			\makebox[\linewidth]{\resizebox{\linewidth}{!}{%
					\begin{tabular}{p{0.2\textwidth}<{\centering}l}
						\toprule
						Original Sentence & 回眸一笑\enspace 就\enspace 好\enspace It is enough to look back and smile \\
						\midrule
						STB (with Cyc) & 回眸一笑\enspace 就\enspace 好\enspace 了\enspace It would be just fine to look back and smile \\
						Ours & 回眸一笑\enspace ,\enspace 勿念\enspace 。\enspace Look back and smile, please do not miss me.
						\\
						\toprule
						Original Sentence & 得过且过\enspace 吧\enspace !\enspace Just live with it!
						\\
						\midrule
						STB (with Cyc) & 想不开\enspace 吧\enspace ,\enspace 我\enspace 的\enspace 吧\enspace 。\enspace I just take things too hard. *
						\\
						Ours & 爱到深处\enspace ,\enspace 随遇而安\enspace 。\enspace Love to the depths, enjoy myself wherever I am.
						\\
						\toprule
						Original Sentence & 自己\enspace 的\enspace 幸福\enspace 给\enspace 别人\enspace 了\enspace Give up your happiness to others
						\\
						\midrule
						STB (with Cyc) & 自己\enspace 的\enspace 幸福\enspace 给\enspace 别人\enspace ,\enspace 你\enspace 的\enspace 。\enspace Give up your happiness to others. *
						\\
						Ours & 自己\enspace 的\enspace 幸福\enspace 是\enspace 自己\enspace ,\enspace 自己\enspace 的\enspace 。\enspace Leave some happiness to yourself, yourself.
						\\
						\bottomrule
			\end{tabular}}}	
			\caption{\label{tab:chat_example}Example sentences on Chat transferred into a romantic style. English translations are provided (* denotes that the sentence has grammar mistakes in Chinese).}
		\end{center}
		\vskip -.25in
	\end{table*}
\end{CJK*}

\end{document}